\documentclass[conference]{IEEEtran} 
\usepackage{cite}
\usepackage{amsmath,amssymb,amsfonts}
\usepackage{algorithmic}
\usepackage{graphicx}
\usepackage{textcomp}
\usepackage{xcolor}
\usepackage{subfigure}
\usepackage{multirow}
\usepackage{multicol}
\usepackage{booktabs}
\def\BibTeX{{\rm B\kern-.05em{\sc i\kern-.025em b}\kern-.08em
    T\kern-.1667em\lower.7ex\hbox{E}\kern-.125emX}}

\graphicspath{{figures/}}

\begin{document}

\title{DRAL: Deep Reinforcement Adaptive Learning for Multi-UAVs Navigation in Unknown Indoor Environment}

\author{Kangtong Mo$^{1,*}$, Linyue Chu$^{1,a}$, Xingyu Zhang$^{2,a}$ Xiran Su$^{2,b}$, Yang Qian$^3$,Yining Ou$^4$ and Wian Pretorius$^5$\\
$^{1,*}$University of Illinois Urbana-Champaign, IL 61820, USA\\
$^{1,a}$University of California, Irvine,  CA 92697, USA\\
$^{2,a}$George Washington University, DC 20052, USA\\
$^{2,b}$Beijing Sineva Robot Technology Co., Ltd at Beijing, 100176, China\\
$^3$University of Southern California, CA 95035, USA\\
$^4$Santa Clara University, CA 95053, USA\\
$^5$Mangosuthu University of Technology, Umlazi 4026, South Africa\\
\centerline{$^{1,*}$mokangtong@gmail.com, $^{1,a}$linyuec@uci.edu}\\ \centerline{$^{2,a}$xingyu\_zhang@gwmail.gwu.edu, $^{2,b}$susirian@gmail.com, $^3$yqian442@usc.edu, $^4$you@scu.edu,$^5$wprs81@gmail.com}
}

\maketitle

%%%%%%%%% ABSTRACT
\begin{abstract}
Autonomous indoor navigation of UAVs presents numerous challenges, primarily due to the limited precision of GPS in enclosed environments. Additionally, UAVs' limited capacity to carry heavy or power-intensive sensors, such as overheight packages, exacerbates the difficulty of achieving autonomous navigation indoors. This paper introduces an advanced system in which a drone autonomously navigates indoor spaces to locate a specific target, such as an unknown Amazon package, using only a single camera. Employing a deep learning approach, a deep reinforcement adaptive learning algorithm is trained to develop a control strategy that emulates the decision-making process of an expert pilot. We demonstrate the efficacy of our system through real-time simulations conducted in various indoor settings. We apply multiple visualization techniques to gain deeper insights into our trained network. Furthermore, we extend our approach to include an adaptive control algorithm for coordinating multiple drones to lift an object in an indoor environment collaboratively. Integrating our DRAL algorithm enables multiple UAVs to learn optimal control strategies that adapt to dynamic conditions and uncertainties.\\
\end{abstract} 


\section{Introduction}
\label{sec:intro}

Micro Aerial Vehicles (MAVs), such as quadcopters outfitted with cameras, have become integral to a wide range of applications~\cite{zhou2024reconstruction,ni2024earnings,jiang2024advanced,li2023mimonet,article,jiang2024dog,kang20216}, including search and rescue operations, exploratory missions, and entertainment. While outdoor autonomous navigation has achieved significant success leveraging the global positioning system (GPS), the precision of GPS diminishes substantially in indoor environments, posing substantial challenges for autonomous indoor flight.

\begin{figure}[!t]
\centering
\includegraphics[width=0.97\linewidth]{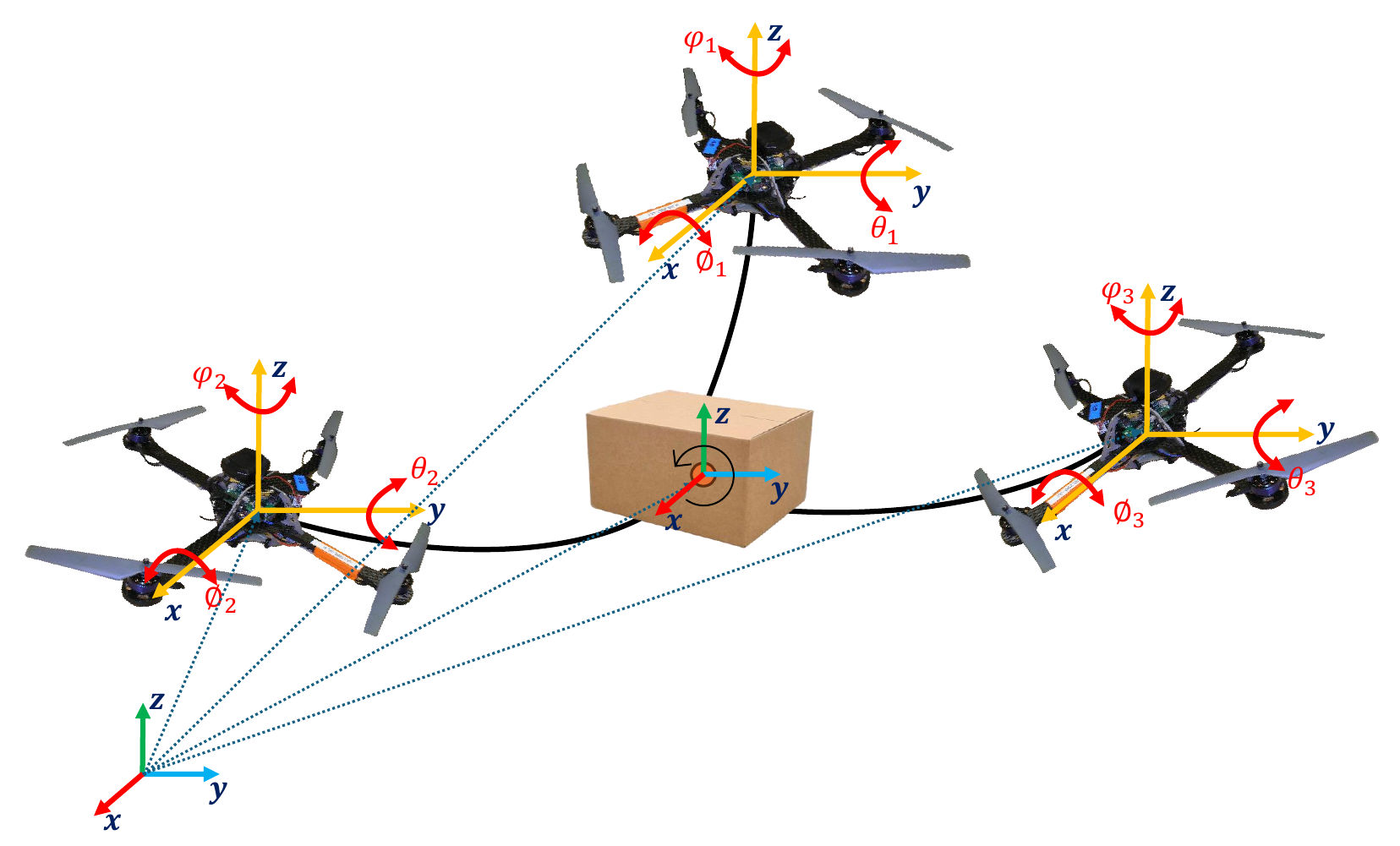}
 \caption{A multi-drone system using cables to lift an unknown payload to transport it during the indoor navigation.}
 \label{fig:overview}

\end{figure}

Various methodologies have been proposed to address the complexities of indoor autonomous navigation. One prominent approach is Simultaneous Localization and Mapping (SLAM), which employs laser range finders, RGB-D sensors~\cite{li2024genie}, or single cameras to generate a three-dimensional map of unknown indoor environments and ascertain the MAV's position within this map. Another notable method~\cite{nikkhoo2023pimbot} involves stereo vision, wherein depth perception is achieved by calculating the disparity between stereo images. However, SLAM is often impractical for MAVs due to the high computational demand of constructing a 3D model~\cite{sun2023instance}, and the resultant 3D structures tend to underperform in feature-sparse environments such as plain walls. Similarly, depth estimation via stereo vision suffers from poor performance in texture-less regions and issues with specular reflections. Additionally, the prevalence of commercially available quadcopters equipped with only a single camera renders these solutions impractical for widespread applications~\cite{huang2024ar,zhu2023demonstration}.

Modern environments are primarily designed for human use, making humanoid robots with their human-like skeletal structures particularly well-suited for tasks in these settings. \cite{liu2024enhanced}, a novel method with the SVM model has been proposed to pave the path to solving the issues in this field, which shows good performance in complicated and changing scenarios. This method offer distinct advantages over other types of robots in human-centric environments. Recently, massively parallel deep reinforcement learning~\cite{jiang2022quo,lu2024cats,DBLP:conf/acl/ChenCL0LT023} in simulation has gained popularity. However, due to the intricate structure of humanoid robots, the sim-to-real gap is more pronounced compared to that of quadrupedal robots. One critical challenge in achieving synchronization within drone networks stems from the inevitable variations in their dynamics and parameters~\cite{choudhury2022dynamic}. For this problem, ~\cite{Gao2024DecentralizedAA} presents a novel decentralized adaptive control method that enables a team of drones to cooperatively transport unknown payloads, effectively adapting to uncertainties in mass, inertia, and grasping points in both gravitational and zero-gravity environments, demonstrating robustness even in the event of vehicle loss during the mission. Also, an extension work in~\cite{gao2024adaptive} makes a significant contribution to aerospace robotics by introducing, for the first time, a novel adaptive detumbling algorithm specifically designed for a non-rigid satellite with unknown dynamic properties, offering a pioneering and high-efficiency solution to address the complexities of flexible satellite stabilization in space, a challenge that had not been fully tackled in previous research. These discrepancies arise from manufacturing inconsistencies, wear and tear, environmental factors~\cite{LI2021101936,dang2024realtime,jiang2023enhancing}, and differences in payloads, presenting substantial obstacles to developing effective control strategies for the network as~\cite{gao2023autonomous}. The disparity between the behaviors exhibited by agents in simulation versus those observed in real-world scenarios is referred to as the reality gap.

To address this, we introduce a deep reinforcement adaptive learning framework designed to train locomotion skills for multiple drones, focusing on zero-shot transfer from simulation to real-world environments. our method incorporates specialized reward structures and domain randomization techniques tailored for multiple drones, thereby mitigating the challenges associated with sim-to-real transfer. 

This paper explores the integration of state-of-the-art (SOTA) adaptive control with deep reinforcement learning methods~\cite{tao2023mlad,sun2023manifold,kang2022tie,tao2023sqba,yan2022influencing} for multi-drone systems, specifically targeting navigation tasks within indoor environments. The contributions of the paper are:
\begin{itemize}
    \item We develop a deep reinforcement adaptive learning algorithm for multiple drones to carry an unknown payload using cables to navigate in an indoor environment.
    \item Our method can ensure the multi-UAV system can carry the payload without prior knowledge of the payload, such as mass, inertia, shape, and so on, to ensure strong extensibility for any payload transportation.
    \item Our algorithm has been proven through various payload and indoor environments in simulations to ensure stability and adaptability.
\end{itemize}

\section{Related work}

Robotic exploration, which leverages multi-robot systems to navigate and map uncharted environments, has been the subject of extensive research. Some studies prioritize rapid spatial exploration~\cite{zhang2024deepgi}, aligning with the objectives of this paper, while~\cite{ni2024timeseries} emphasize precise environmental reconstruction. Among the myriad methodologies proposed, frontier-based approaches stand out as classic techniques. Moreover, \cite{zhang2024developmentapplicationmontecarlo} developed a novel algorithm for a Monte Carlo tree search algorithm, which improved the searching efficiency for robotics decision tasks. These methods were initially introduced in related works and subsequently subjected to a more thorough evaluation in \cite{wang2024research}. The closest frontier was chosen as the next target in the seminal work.
Conversely,~\cite{zhang2020manipulator} presented an alternative strategy that, during each decision-making process, selects the frontier within the field of view that minimizes velocity changes, thereby maintaining a consistently high flight speed. This strategy has been demonstrated to outperform the traditional method~\cite{liu2024td3,zhu2021twitter,hu2023artificial,zhu2022optimizing}. Furthermore, in~\cite{afram2014theory}, a differentiable measure of information gain based on frontiers was introduced, enabling path optimization through gradient information.

Utilizing range sensors or visual sensors, a three-dimensional mapping of unexplored indoor environments can be constructed while concurrently estimating the sensor's position within the map.~\cite{li2019segmentation} implemented a high-level SLAM system leveraging a laser rangefinder to navigate and map uncharted indoor spaces. Similarly,~\cite{li2023deception} proposed a monocular vision-based SLAM approach for autonomous indoor navigation. Nevertheless, due to the 3-D reconstruction process, SLAM's computational intensity results in significant latency between perception and action. Additionally, SLAM's accuracy diminishes in indoor settings with insufficient feature points for consistent frame-to-frame tracking, such as blank walls. Our system circumvents the need for path-planning, thereby prioritizing rapid responsiveness to the immediate environment. Consequently, it demonstrates robust performance in detecting and avoiding obstacles like walls, effectively minimizing delays.

Most extant methodologies tend to employ greedy decision-making processes~\cite{lipeng2024prioritized}, neglecting the multi-drone system's and payload's inherent dynamics. This oversight results in suboptimal global tours and overly cautious maneuvers. In stark contrast, our approach devises tours that proficiently encompass the entire environment while concurrently generating dynamically feasible~\cite{zhang2024enhancing}, minimum-time trajectories for multi-drone systems in unknown payload transportation tasks in an indoor environment. This enables the multi-drone system to execute agile and efficient flights, maximizing both coverage and performance.

\section{Methodology}
\subsection{Multi-drone Dynamics}
The proposed collaborative system consists of hex-rotors attached to a payload as shown in Fig.~\ref{fig:overview}. The payload is a rigid length $d$ and diameter $w$ cylinder with uniformly distributed mass. Examples of such payloads in the real world include wooden logs, pre-cast columns, and pipeline segments. In real-world scenarios, this payload could be any item such as rechargeable batteries, communication equipment, beams, fire fighting equipment, sensors like or spray for agricultural purposes. We define the position vector $P_i=[x, y, z]^\intercal$ of the CoM (Center of mass) of the slung load relative to a fixed inertial frame $\varepsilon=\left[x^{\prime}, y^{\prime}, z^{\prime}\right]^\intercal$. We define the orientation of the slung load by using
Euler angles as $\zeta=[\alpha, \beta, \gamma]^\intercal$, where $\alpha$ represents the roll angle about the x-axis which is
described as tilt angle. The angle $\alpha$ is the main constituent in the load distribution strategy.

When considered as a whole, the system consists of a combined mass, UAVs as actuators, facing parasitic drag forces caused by air friction. Extending on the single UAV dynamic model, the lumped dual-UAV-payload dynamics can be defined as:
\begin{equation}
\begin{aligned}
m \ddot{x} & =\sum_{i=1}^3\left(\sin \phi_i \sin \psi_i+\cos \phi_i \cos \psi_i \sin \theta_i\right) S_i \\
& +F_{D i}\left(V_x+V_{w x}(z), \theta_i, \rho(z)\right)+F_{p x}
\end{aligned}
\end{equation}

\begin{equation}
\begin{gathered}
m \ddot{y}=\sum_{i=1}^2\left(\cos \phi_i \sin \theta_i \sin \psi_i-\cos \psi_i \sin \phi_i\right) S_i \\
+F_{D i}\left(V_y+V_{w y}(z), \phi_i, \rho(z)\right)+F_{p y} \\
m \ddot{Z}=\sum_{i=1}^2\left(\cos \theta_i \cos \phi_i\right) T_i-m g \\
+F_{D i}\left(V_z+V_{w y}(z), \phi_i, \rho(z)\right)+F_{p z} .
\end{gathered}
\end{equation}
where $m=m_p+\sum_{i=1}^2 M_i$ is the total mass which includes mass of hexrotors $M_i$ and
slung load $m
_p$, $\ddot{x}, \ddot{y}, \ddot{z}$ are the translational accelerations of the dual-hexrotor-UAV payload system in $x$, $y$ and $z$ axes. $S_i$ is the sum of the thrusts produced by all motors of the UAV $T_i=k_b \sum_{N=1}^6 \omega_{i}^2$. Payload’s drag forces are shown by $F_{p x}, F_{p y}, F_{p z}$ for $x,y,z$
axis respectively.

\subsection{Deep reinforcement adaptive learning Design}

Our methodology leverages a sophisticated reinforcement learning paradigm $\mathcal{M} = \langle \mathcal{S}, \mathcal{A}, T, \mathcal{O}, R, \gamma \rangle$, where $\mathcal{S}$ and $\mathcal{A}$ signify the state and action spaces, respectively. The transition dynamics are represented by $T(\mathbf{s}'|\mathbf{s},\mathbf{a})$, and the reward function is denoted by $R(\mathbf{s},\mathbf{a})$. The discount factor $\gamma \in [0, 1]$ and the observation space $\mathcal{O}$ are integral components of the framework. This model is adeptly designed for application in both simulated and real-world environments, facilitating a shift from full observability in simulations $(\mathbf{s}\in\mathcal{S})$ to partial observability in real-world scenarios ($\mathbf{o}\in\mathcal{O}$). This transition necessitates the utilization of a Partially Observable Markov Decision Process, wherein the policy $\pi(\mathbf{a}|\mathbf{o}_{\leq t})$ maps observations to action distributions, thereby optimizing the expected return $J = \mathbb{E}[R_t] = \mathbb{E}\left[\sum_{t}\gamma^\intercal r_t\right]$.

We leverage the Proximal Policy Optimization algorithm \cite{schulman2017proximalpolicyoptimizationalgorithms}, augmented by the Asymmetric Actor Critic approach, incorporating privileged information during the training phase and transitioning to partial observations for deployment. The policy loss is formulated as:
\begin{equation}
\label{eq:policy}
\begin{aligned}
\mathcal{L}_{\pi} &= \min \left[
\frac{\pi(a_t \mid o_{\leq t})}{\pi_{b}(a_t \mid o_{\leq t})} A^{\pi_{b}}(o_{\leq t}, a_t), \right]
\end{aligned}
\end{equation}

Advantage estimation employs Generalized Advantage Estimation, necessitating an updated value function:

\begin{equation}
\label{eq:value}
\mathcal{L}_v = \| R_t - V(s_t)\|_2,
\end{equation}

\section{Experiment Results}

\begin{figure}[!t]
\centering
\includegraphics[width=0.97\linewidth]{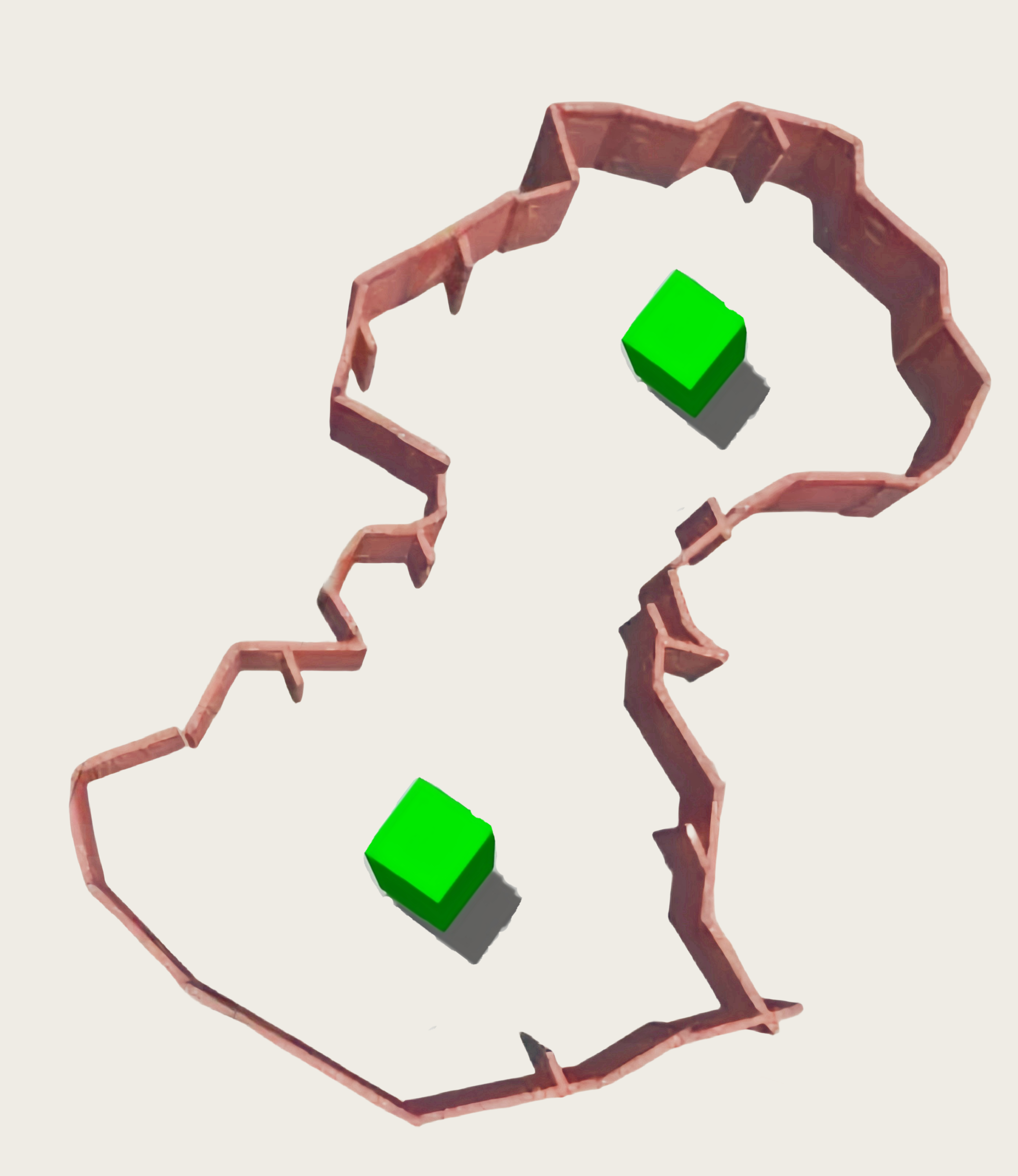}
 \caption{Simulation environment setup in Gazebo.}
 \label{fig:2}

\end{figure}

\begin{figure}[!t]
\centering
\includegraphics[width=0.97\linewidth]{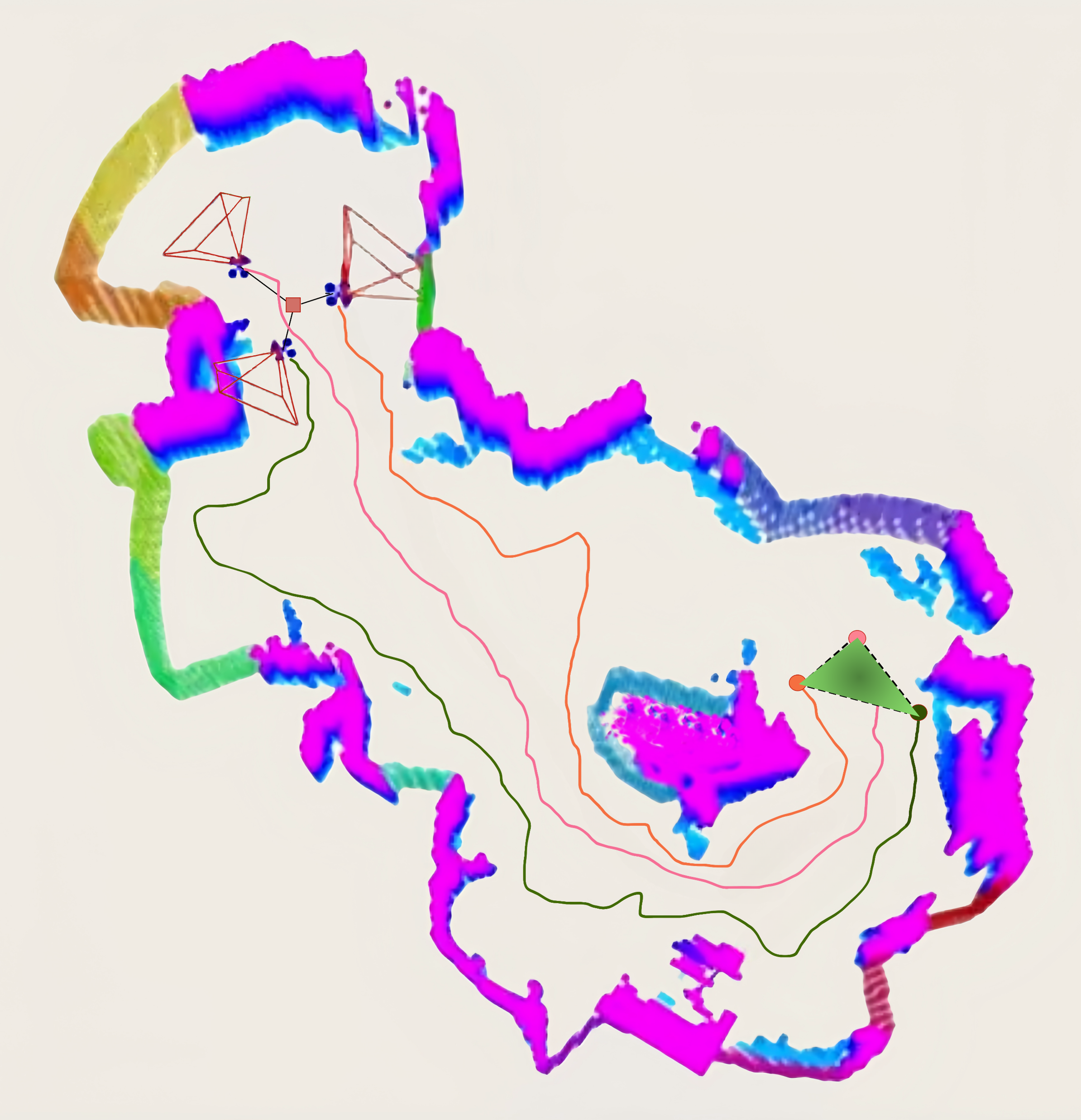}
 \caption{Three quadrotors elevate an unidentified payload during an autonomous exploration assessment executed within a multifaceted indoor environment.}
 \label{fig:3}

\end{figure}

In this section, we delineate the simulated and empirical evaluation methodologies. The results derived from these assessments substantiate our hypothesis that our DRAL method is capable of attaining superior performance in intricate and uncharted environments.

In our experimental setup, we employ three drones to elevate an unidentified box within a simulation environment modeled on an indoor room constructed in Gazebo, as illustrated in Fig.~\ref{fig:2}. The local grid maps encompass an area of  $14m \times 8m$. Each drone is stationed at the center of its respective local map and is confined to perceiving information within a 3-meter radius, consequently constraining its long-range planning capabilities as shown in Fig.~\ref{fig:3}. These parameters are uniformly maintained across both simulated and real-world experimental conditions.

\begin{figure}[!t]
\centering
\includegraphics[width=0.97\linewidth]{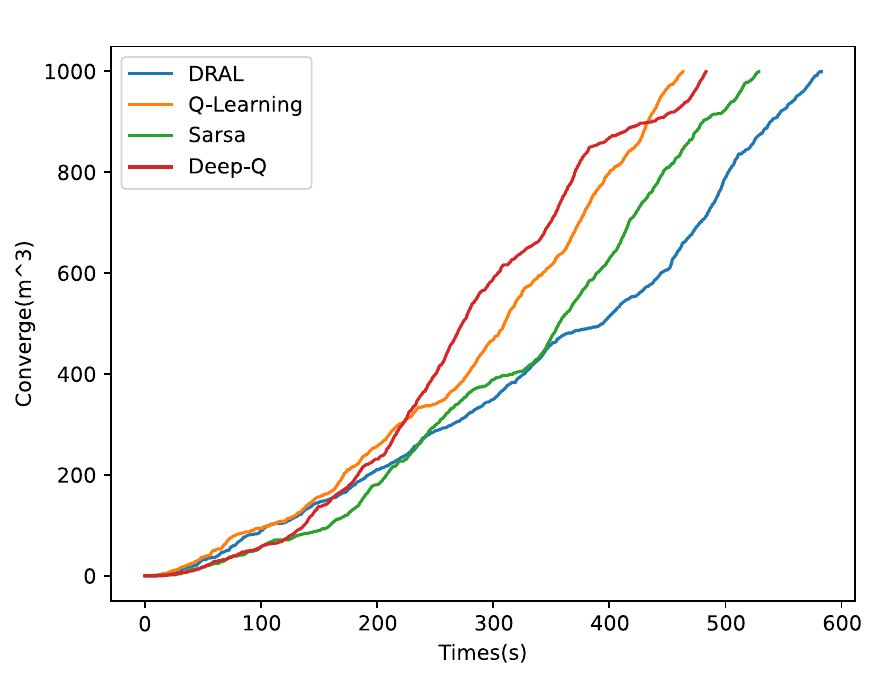}
 \caption{The investigative advancement of four methodologies. The results show that our method is better than the other three popular deep learning methods.}
 \label{fig:4}

\end{figure}

We evaluate our proposed framework through simulation in Gazabo, benchmarking it in Q-learning, Deep Q learning Sarsa and our method to compare each other. In all tests, the dynamic limits are uniformly set to  $v_{\text{max}} = 3.4$  m/s and  $\omega_{\text{max}} = 0.57$  rad/s for all methods. The sensors' fields of view are configured to  $[50 \times 50] \ \text{deg}$  with a maximum range of 3 meters. Each method is executed three times in both scenarios, maintaining the same initial configuration. The four methods' statistical outcomes and exploration progress are presented in Tab.~\ref{tab:result} and Fig. \ref{fig:4}, respectively. Tab.~\ref{tab:result} shows that our method shows a higher success rate than the other three popular methods in the different transportation tasks during indoor navigation, which means the success rate for three drones carrying an unknown payload can transport from the initial position to the target position without any collision and crush is higher than other methods' verifications.

The results elucidate that our approach manifests substantially abbreviated exploration durations and diminished temporal variance. The aggregate exploration trajectory delineated by our method is markedly more concise, predominantly attributable to our global tour planning strategy. The executed trajectory exhibits enhanced smoothness, a consequence of our local motion refinement and the generation of fluid trajectories. Furthermore, our minimum-time trajectory planning algorithm facilitates our capacity for navigation at elevated flight velocities.

	\begin{table}[!tp]
		\centering
		\setlength\tabcolsep{3.5 pt}
		\caption{Performace comparison of different Algorithm: we choose different classes of payload in the same simulation environment to test our method with other different methods.}
		\begin{tabular}{rlrrrrrr}
			\toprule
			\multicolumn{1}{l}{Metric} & Method & \multicolumn{2}{l}{Object Class} &       &       &        \\
			\cmidrule{3-7}          &       & Box & Package & Bucket &  \\
			\midrule
			\multicolumn{1}{l}{Success} & Q Learning   & 0.502 & 0.517 & 0.612          \\
			\multicolumn{1}{l}{rate} & Deep-Q Learning & 0.614 & 0.603 & 0.706          \\
			\multicolumn{1}{l}{}& Sarsa & 0.622 & 0.618 & 0.786 \\
			& \textbf{DRAL} & \textbf{0.824} & \textbf{0.857} & \textbf{0.903} \\
			\midrule
			\multicolumn{1}{l}{Reach} & Q Learning   & 52.4  & 56.7  & 48.2          \\
			\multicolumn{1}{l}{time} & Deep-Q Learning & 47.3  & 49.1  & 43.9          \\
			\multicolumn{1}{l}{(s)}& Sarsa & 39.4  & 42.5  & 38.2   \\
			& \textbf{DRAL} & \textbf{24.1}  & \textbf{25.7}  & \textbf{21.9}  \\
			\bottomrule
		\end{tabular}%
		\label{tab:result}%
	\end{table}%

\section{Conclusion}
In this paper, we concentrate on multi-drone systems operating under unknown payloads while navigating in uncharted environments. To enable the robots to adapt to a more diverse distribution of obstacles, we introduce DRAL, a novel deep reinforcement learning method that leverages adaptive control during the unknown payload transportation task with a continuous action space. This method breakthrough the traditional drone navigation problem which only focus on single drone navigation but also consider the payload transportation in 3D space. We demonstrate that DRAL surpasses several baseline schemes across multiple scenarios, both in simulation and real-world settings. Additionally, we conduct ablation experiments to illustrate the efficacy of our individual components.

\bibliographystyle{IEEEtran}
\bibliography{main}

\end{document}